\documentclass{article}
\usepackage{ijcai21}

\usepackage{times}
\usepackage{soul}
\usepackage{url}
\usepackage[hidelinks]{hyperref}
\usepackage[utf8]{inputenc}
\usepackage[small]{caption}
\usepackage{graphicx}
\usepackage{amsmath}
\usepackage{amsthm}
\usepackage{booktabs}
\usepackage{algorithm}
\usepackage{algorithmic}
\usepackage{subfigure,amsfonts,multicol,pdfpages}
\title{Rethink the Connections\\ among Generalization, Memorization, and the Spectral Bias of DNNs}

\author{
    Xiao~Zhang$^1$\and
    Haoyi~Xiong$^{2}$\and
    Dongrui~Wu$^1$\thanks{Corresponding author.}
    \affiliations
    $^1$ Huazhong University of Science and Technology, Wuhan, China\\
    $^2$ Baidu Research, Bejing, China\\
    \emails
    xiao\_zhang@hust.edu.cn,
    xionghaoyi@baidu.com,
    drwu@hust.edu.cn
}

\begin{document}

\maketitle

\begin{abstract}
Over-parameterized deep neural networks (DNNs) with sufficient capacity to memorize random noise can achieve excellent generalization performance, challenging the bias-variance trade-off in classical learning theory. Recent studies claimed that DNNs first learn simple patterns and then memorize noise; some other works showed a phenomenon that DNNs have a spectral bias to learn target functions from low to high frequencies during training. However, we show that the monotonicity of the learning bias does not always hold: under the experimental setup of deep double descent, the high-frequency components of DNNs diminish in the late stage of training, leading to the second descent of the test error. Besides, we find that the spectrum of DNNs can be applied to indicating the second descent of the test error, even though it is calculated from the training set only.
\end{abstract}

\section{Introduction}

The bias-variance trade-off in classical learning theory suggests that models with large capacity to minimize the empirical risk to almost zero usually yield poor generalization performance. However, this is not the case of modern deep neural networks (DNNs): Zhang \emph{et al.}~\shortcite{zhang2017} showed that over-parameterized networks have powerful expressivity to completely memorize all training examples with random labels, yet they can still generalize well on normal examples. This phenomenon cannot be explained by the VC dimension or Rademacher complexity theory.

Some studies attributed this counterintuitive phenomenon to the implicit learning bias of DNN's training procedure: despite of the large hypothesis class of DNNs, stochastic gradient descent (SGD) has an inductive bias to search the hypothesises which show excellent generalization performances. Arpit \emph{et al.}~\shortcite{Arpit2017} claimed that DNNs learn patterns first, and then use brute-force memorization to fit the noise hard to generalize. Using mutual information between DNNs and linear models, Kalimeris \emph{et al.}~\shortcite{Kalimeris2019} showed that SGD on DNNs learns functions of increasing complexity gradually. Furthermore, some studies showed that lower frequencies in the input space are learned first and then the higher ones, which is known as the spectral bias or frequency principle of DNNs~\cite{Rahaman2019,Xu2019,Xu2019a}. They showed that overfitting happens when the complexity of models keeps increasing or high-frequency components remain being introduced.

All the findings above are based on a basic assumption that the learning bias in training DNNs is monotonic, e.g., from simple to complex or from low frequencies to high frequencies. However, the monotonicity of the training procedure was recently challenged by epoch-wise double descent: the generalization error first has a classical U-shaped curve and then follows a second descent~\cite{Nakkiran2020}. It is intriguing because according to spectral bias, with higher-frequency components being gradually introduced in training, the generalization performance should deteriorate monotonically due to the memorization of noise.

\begin{figure}[ht]\centering
\includegraphics[width=.85\columnwidth,clip]{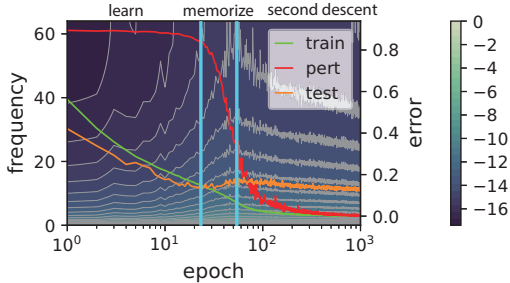}
\caption{Heat map with contour lines showing the energy ratio (in logarithmic scale) of different frequency components at every epoch, synchronized with training, perturbed, and test errors. The model was ResNet18 trained on CIFAR10 with 10\% label noise (Adam with learning-rate 1e-4 for 1,000 epochs). The horizontal axis is shown in logarithmic scale. The vertical lines indicate early stopping and the start of the second descent, dividing the training procedure into three phases.}
\label{fig:intro}
\end{figure}

To better understand the connections among generalization, memorization and the spectral bias of DNNs, we explored the frequency components of learned functions under the experimental setup of double descent (randomly shuffle the labels of part of the training set and train DNNs for an extended number of epochs). We surprisingly observed that at a certain epoch, usually around the start of the second descent, while the perturbed part is still being memorized, the high-frequency components begin to diminish (see the \emph{second descent} phase in Figure~\ref{fig:intro}). After exploring this phenomenon in a traceable toy task, we show that it happens because the prediction surface off the training data manifold becomes flatter and more regularized in the late training stage, which improves the generalization performance of models on the test points that are not covered by the training data manifold. As a result, the second descent of the test error happens.

We further show that though it does not change monotonically, the spectrum manifests itself as an indicator of the training procedure. We trained different DNNs on some image classification datasets, verifying the connections between the second descent of the test error and the diminishment of the high-frequency components, even if the spectrum is calculated from the training set. It suggests that monitoring the test behaviors with only the training set is possible, which provides a novel perspective to studying the generalization and memorization of DNNs in both theory and practice.

The remainder of the paper is organized as follows: we first present our analysis of the spectra of DNNs, then we show that it is possible to monitor the test behaviors according to the spectrum. The following section summarizes some related works. The last section draws conclusions.

\paragraph{Contributions.}
Our major contributions are highlighted as follows:
\begin{itemize}
\item We provide empirical evidence to show that the monotonicity of the learning bias does not always hold. In the late stage of training, though high-frequency components are introduced to memorize the label noise, the prediction surface off the training data manifold is more biased towards low-frequency components, leading to the second descent of the test error.
\item We find that unlike errors, the spectra of DNNs vary consistently on the training sets and test sets.
\item We explore the test curve and the spectrum calculated on the training set, and discover a correlation between the peak of the spectrum and the start of the second descent of the test error, which suggests that it may be possible to monitor the test behaviors using the training set only.
\end{itemize}
\section{Frequency Components in Training}

In this section, we analyze the spectra of DNNs trained on several image classification benchmarks, and find that the monotonicity of the spectral bias does not always hold. Moreover, we show that the diminishment of the high-frequency components leads to the second descent of the test error.

\subsection{Fourier Spectrum}

Fourier transforming on DNNs can be a tough task due to the high dimensionality of the input space. Rahaman \emph{et al.}~\shortcite{Rahaman2019} exploited the continuous piecewise-linear structure of ReLU networks to rigorously evaluate the Fourier spectrum, but their approach cannot be performed on DNNs trained on high-dimensional datasets. Xu \emph{et al.}~\shortcite{Xu2019} used non-uniform discrete Fourier transform (DFT) to capture the global spectrum of a dataset, but it cannot be very accurate due to the sparsity of points in high-dimensional space.

In this paper, we propose a heuristic but more practical metric to measure the spectrum of a DNN. Instead of capturing the frequency components of the whole input space, we pay more attention to the variations of the DNN in local areas around data points. We shall denote the input point sampled from a distribution $\mathcal{D}$ by $\boldsymbol{x} \sim \mathcal{D}$, a normalized random direction by $\boldsymbol{v}_{x}$, the $c$-th logit output of a DNN by $f_c(\boldsymbol{x})$, where $c\in\{1, 2, ..., C\}$ and $C$ is the number of classes. We evenly sample $N$ points from $[\boldsymbol{x} - h\boldsymbol{v}_x, \boldsymbol{x} + h\boldsymbol{v}_x]$ to perform the discrete Fourier transform, where $h$ bounds the area. The Fourier transform of $f_c(\boldsymbol{x})$ is then:
\begin{align}
    \tilde{f}_{c,\boldsymbol{x}}(k)=\sum_{n=1}^{N}f_c\left(\boldsymbol{x}+\frac{2n-N-1}{N-1}
    h\boldsymbol{v}_x\right)e^{-i2\pi\frac{n}{N}k}. \label{eq:fck}
\end{align}
We use the logit outputs instead of the probabilities so that the spectrum is irrelevant to the rescaling of the DNN parameters. Because, when the weights of the last layer are multiplied by $\alpha>0$ (this operation does not change the decision boundary), the spectrum will change nonlinearly if we use the probability outputs. We then add up the spectra across the dataset and the logit outputs to illustrate the local variation from a global viewpoint:
\begin{align}
    A_k=\frac{1}{C}\sum_{c=1}^{C}\mathbb{E}_{\boldsymbol{x}\sim\mathcal{D}}\left|\tilde{f}_{c,\boldsymbol{x}}(k)\right|^2. \label{eq:Ak}
\end{align}

In practice, we only need a small number (as tested in our experiments, $500$ data points are usually enough) of data points to approximate the expectation of $\left|\tilde{f}_{c,\boldsymbol{x}}(k)\right|^2$ in (\ref{eq:Ak}).

Note that computing $A_k$ does not need any label information of the dataset, suggesting that $A_k$ may be applied to semi-supervised or unsupervised learning.

\begin{figure*}[ht]\centering
\subfigure[SVHN]{\label{fig:svhn}
    \begin{minipage}[b]{0.32\textwidth}
    \includegraphics[width=\textwidth]{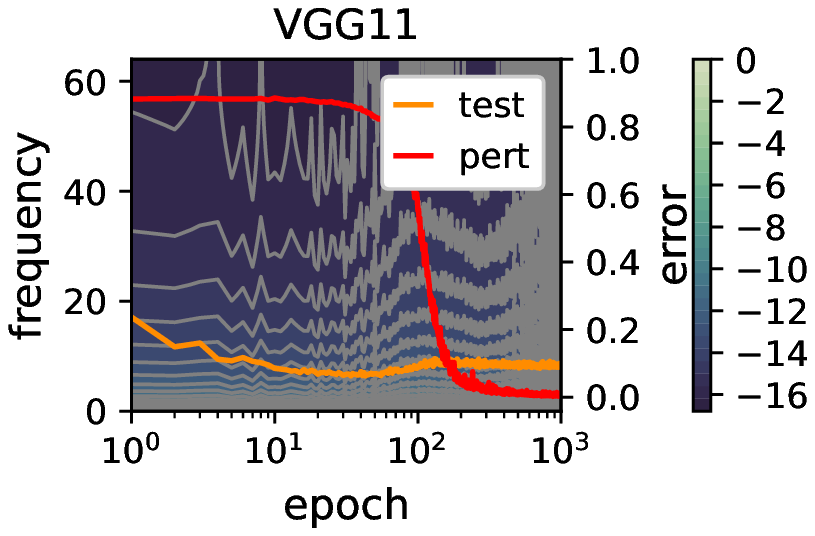} \\
    \includegraphics[width=\textwidth]{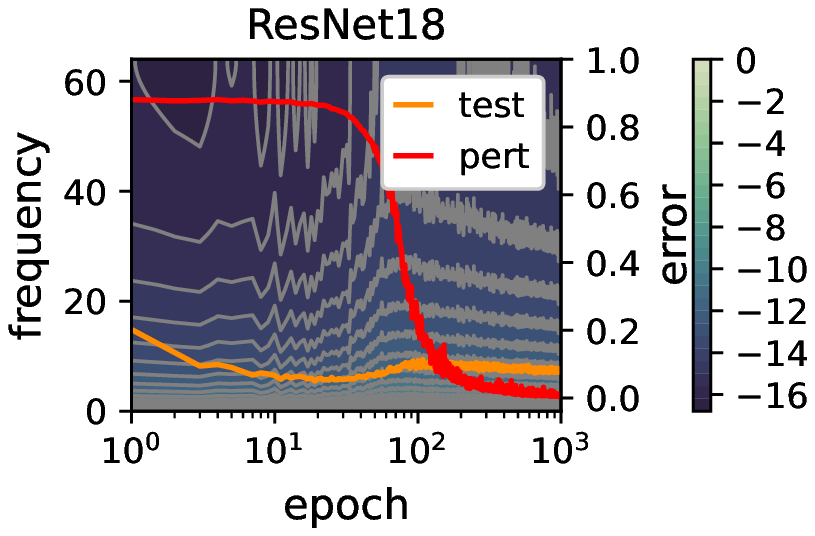}
    \end{minipage}
}
\subfigure[CIFAR10]{\label{fig:cifar10}
    \begin{minipage}[b]{0.32\textwidth}
    \includegraphics[width=\textwidth]{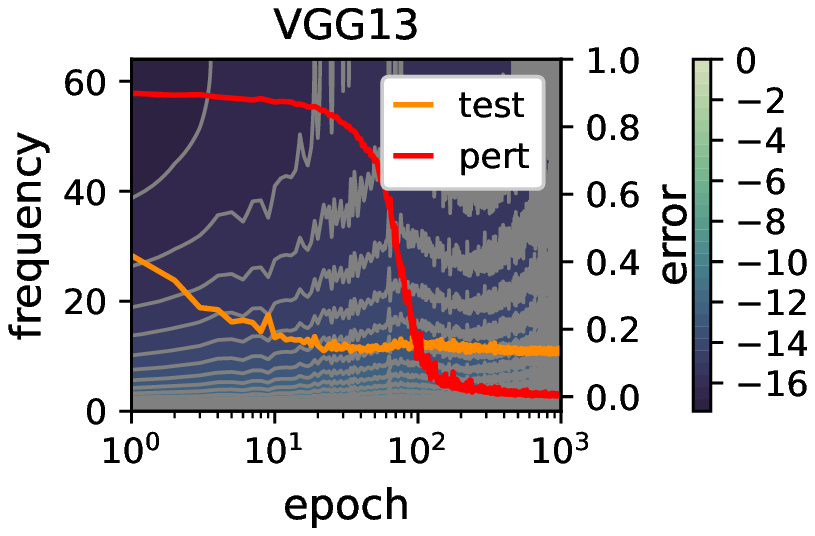} \\
    \includegraphics[width=\textwidth]{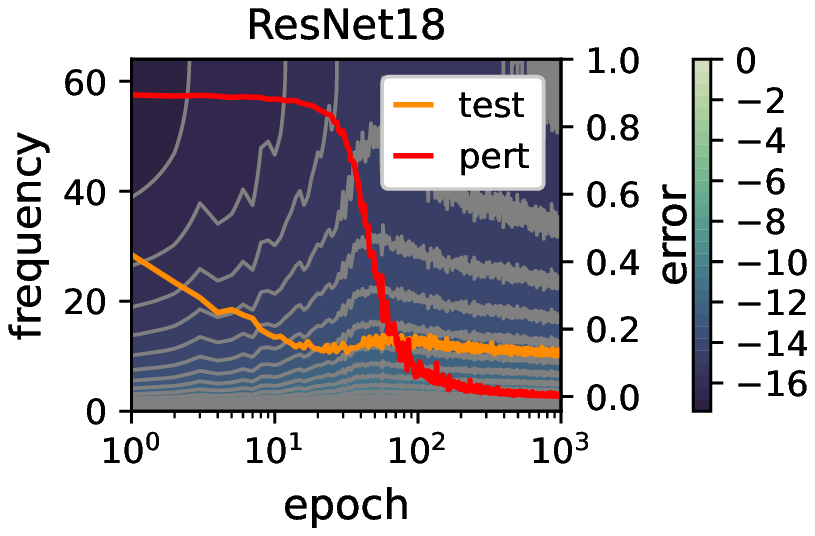}
    \end{minipage}
}
\subfigure[CIFAR100]{\label{fig:cifar100}
    \begin{minipage}[b]{0.32\textwidth}
    \includegraphics[width=\textwidth]{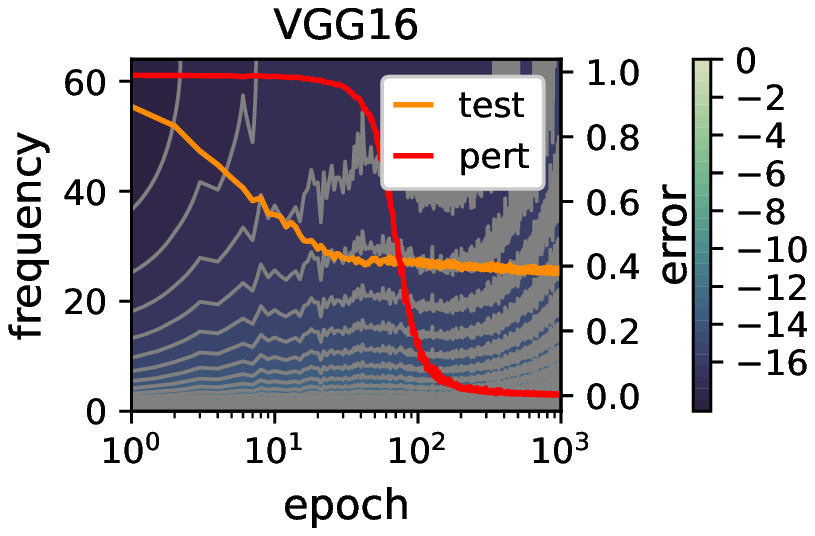} \\
    \includegraphics[width=\textwidth]{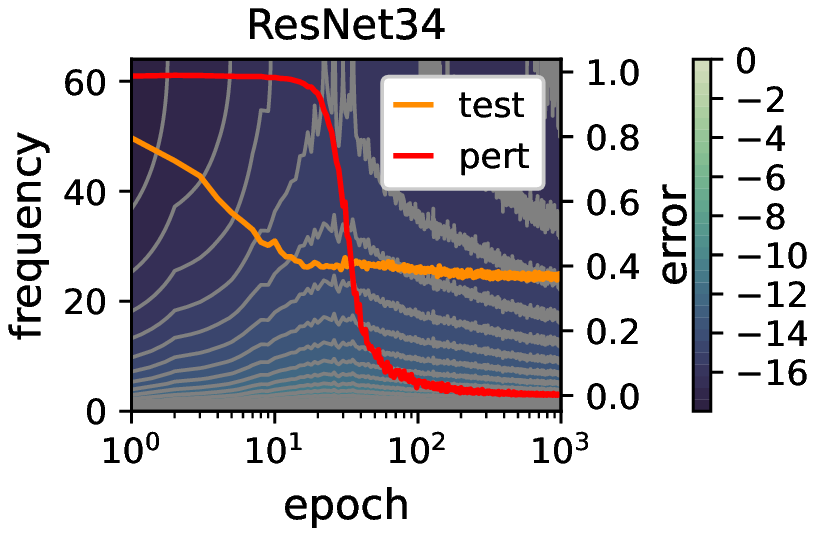}
    \end{minipage}
}
\caption{ResNet and VGG trained on SVHN, CIFAR10 and CIFAR100 with 10\% label noise. The red curves indicate the errors on the perturbed set, whereas the orange ones indicate the errors on the test set. The heat map with contour lines depicts the spectrum $R_k$, which is calculated on the training set. The horizontal axis is shown in logarithmic scale. Appendix~A presents results of other levels of label noise.} \label{fig:FreqAndError}
\end{figure*}

\subsection{Non-Monotonicity of the Spectral Bias} \label{sect:spec}

Previous studies on spectral bias assumed that its evolutionary process is monotonic, i.e., DNNs learn the low frequencies first, and then the high frequencies, which means we should observe a monotonic increase in the ratio of high-frequency components. However, this is not what we observed: we found that at a certain epoch, usually around the epoch when the second descent starts, the ratio of high-frequency components begins to diminish.

Our experimental setup was analogous to deep double descent~\cite{Nakkiran2020}. We here consider two architectures\footnote{Adapted from \url{https://github.com/kuangliu/pytorch-cifar}.} (VGG~\cite{Simonyan2015} and ResNet~\cite{He2016}) on three image datasets (SVHN~\cite{Netzer2011}, CIFAR10 and CIFAR100~\cite{Krizhevsky2009}). We normalized the values of image pixels to $[0,1]$ and randomly shuffled $10\%$ labels of the training set to strengthen double descent, where the perturbed part was denoted by \emph{perturbed set}. The batch-size was set to $128$, and we utilized the Adam optimizer with the learning rate $0.0001$ to train the models for $1,000$ epochs with data augmentation.

To better compare the spectrum along the training process, we calculated the energy ratio of $A_k$ in logarithmic scale:
\begin{align}
    R_k=\log\frac{A_k}{\sum_jA_j}. \label{eq:Rk}
\end{align}
For every epoch, we randomly chose $500$ data points from the training set, sampled a normalized direction $\boldsymbol{v}_{x}$ for each data point, and calculated $R_k$ for the $k$-th frequency component\footnote{We call $k$ ``frequency'' in the rest of the paper.} based on the sampled data points and the directions. In our experiments, we set $h=0.5$ and $N=128$.

The spectra and learning curves are shown in Figure~\ref{fig:FreqAndError}. From the error curves, we can observe that the test error decreases very quickly at the beginning of the training, whereas the error on the perturbed set remains high, suggesting that the models effectively learn the patterns of the data in this period. However, as the training goes on, the error on the perturbed set decreases rapidly, along with a little increase of the test error. In this period, the models start to memorize the noise, which leads to overfitting on the training set. So far, the behaviors of the models are consistent with the conventional wisdom. However, if the models are trained with more epochs, the peak of the test error occurs, just around the epoch when the model memorizes the noise, and then the test error steps into the second descent. This is known as the epoch-wise double descent~\cite{Nakkiran2020}.

The heat map of $R_k$ in Figure~\ref{fig:FreqAndError} presents a new perspective to looking at this phenomenon. The spectral bias manifests itself in the learning and memorization phases: the models first introduce low-frequency components and then the high-frequency ones. The ratio of the high-frequency components increases rapidly when the models try to memorize the noise. Nonetheless, along the second descent of the test error, the high-frequency components begin to diminish, violating the claims about the monotonicity of the spectral bias. The non-monotonicity of the spectral bias implies that limited high-frequency components may be sufficient to memorize the noise, which can be observed during the second descent: though the ratio of the high-frequency components decreases, the perturbed set is still being memorized.

Our experimental results also show that the skip connection significantly influences the spectra. Comparing the late-stage heat maps between VGG and ResNet in Figure~\ref{fig:FreqAndError}, we can observe that ResNet seems to have a bias towards low-frequency components. Since high-frequency components usually suggest complex decision boundary with poor generalization, this may explain why skip connection can improve the performance of DNNs.

\subsection{Why Do High-Frequency Components Diminish?}
Label noise raises a potential demand for the high-frequency components. It is reasonable since the perturbed point resembles the Dirac Delta function on the prediction surface of the training data manifold, which has a broadband power spectrum. However, this explanation may lead to a new question: \emph{why do high-frequency components diminish when the perturbed set is still memorized}?

Notice that the perturbed set remains memorized during the second descent, which means the gain of the generalization performance in this phase is not from fitting the training data manifold but from regularizing the prediction surface in other input space. Inspired by this fact, we consider the spectra along the training data manifold and off the training data manifold separately. We show that the complex variation of the spectrum shown in Section~\ref{sect:spec} is the combination of two processes: the on-manifold prediction surface keeps introducing high-frequency components to fit the perturbed points, whereas the off-manifold prediction surface gradually becomes biased towards low-frequency components.

\begin{figure*}[ht]\centering
\subfigure[]{\label{fig:OnManifold}
    \includegraphics[width=.3\textwidth]{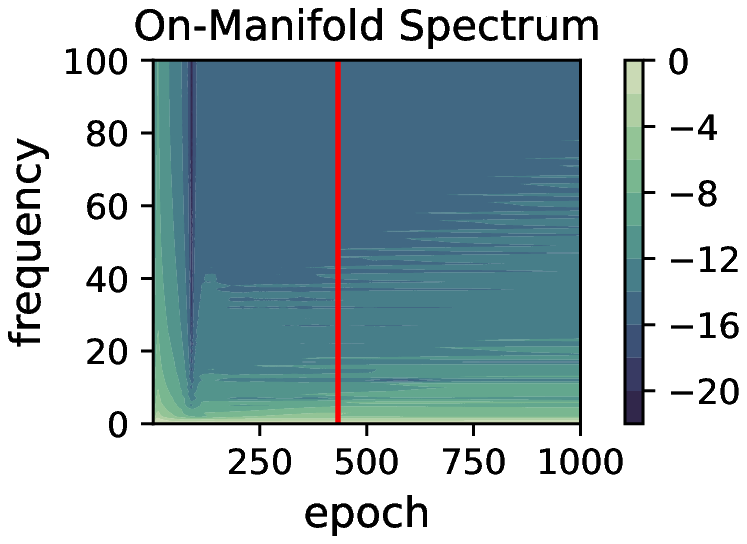}
}
\subfigure[]{\label{fig:OffManifold}
    \includegraphics[width=.3\textwidth]{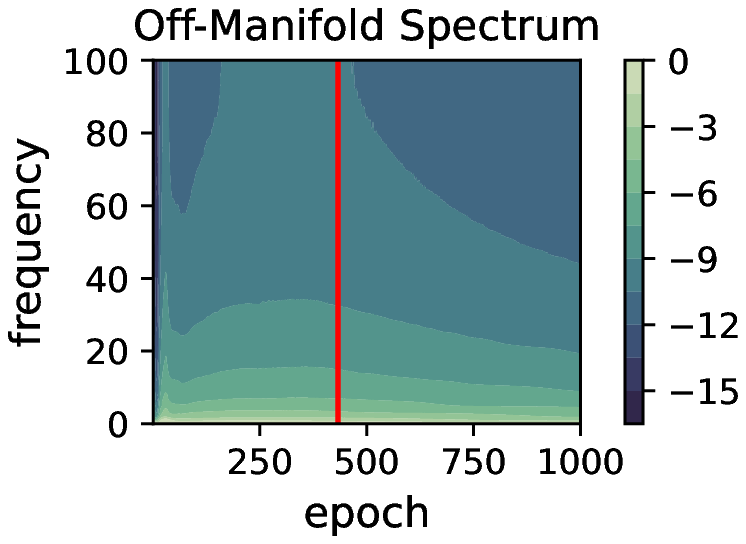}
}
\subfigure[]{\label{fig:OnOffACC}
    \includegraphics[width=.3\textwidth]{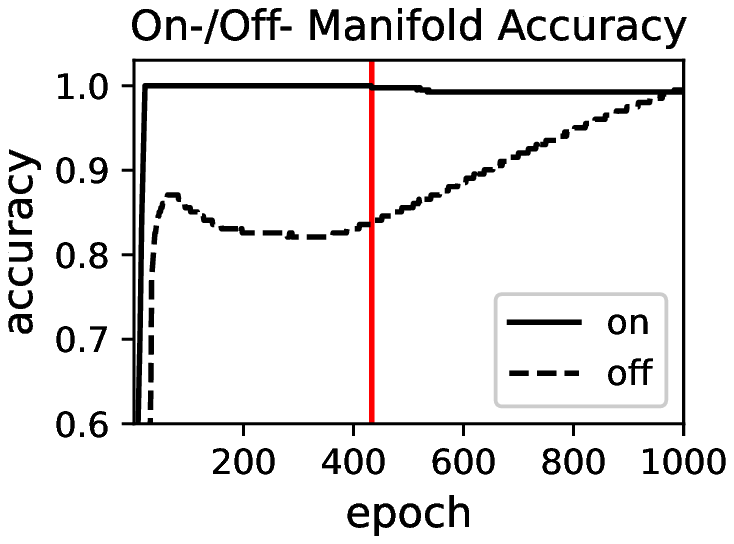}
}
\caption{On-/off- manifold spectra and accuracies. The red lines indicate the epoch when the perturbed point begins to be memorized.} \label{fig:OnOffManifold}
\end{figure*}

To verify our hypothesis, we design a toy task which requires separating two disjoint but perpendicular lines that lie in the three-dimensional space (the label of one data point is perturbed), and then we investigate the variation of the on-/off- manifold spectra and generalization performances. Figure~\ref{fig:OnOffManifold} presents a simple illustration of our toy task.

\begin{figure}[tb]\centering
\includegraphics[width=.9\columnwidth,clip]{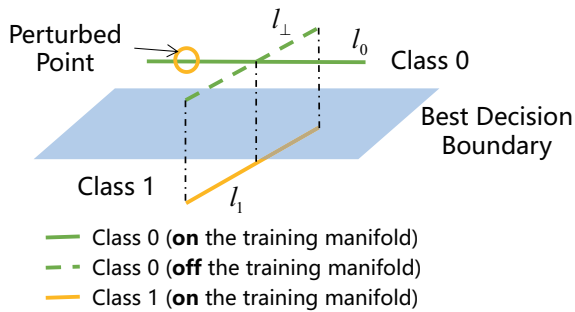}
\caption{Illustration of the toy task. The blue plane is the desired decision boundary separating $l_0$ and $l_1$. The perturbed point is labeled as Class 1. The off-manifold line $l_\perp$ intersects with the on-manifold line $l_0$ perpendicularly. The task is to train the model on $l_0$ and $l_1$, and observe the spectra on $l_0$ and $l_\perp$.} \label{fig:toy}
\end{figure}

Specifically, we define $l_0$ and $l_1$ as:
\begin{align}
l_i=\{\boldsymbol{r}_i+k\cdot \boldsymbol{v}_i|-1\leq k\leq 1\}, i\in\{0,1\},
\end{align}
where $\boldsymbol{r}_0=[0.1, 0.1, 0.1]$, $\boldsymbol{r}_1=-\boldsymbol{r}_0$, $\boldsymbol{v}_0=\frac{1}{\sqrt{2}}[1, -1, 0]$, and $\boldsymbol{v}_1=\frac{1}{\sqrt{6}}[1, 1, -2]$. For the training set, we evenly sample 51 points from $l_0$ and $l_1$, respectively, and label the points in $l_i$ as Class $i$ (except one randomly chosen perturbed point in $l_0$). For the test set, we evenly sample 201 points from $l_0$ and $l_1$, respectively, without any perturbation. A ReLU DNN with two fully-connected hidden layers (100 neurons for each) is applied to train on the training set with Adam optimizer (lr=5e-4). During the training process, we compute both the on-manifold spectrum on $l_0$ and the off-manifold spectrum calculated on $l_{\perp}=\{\boldsymbol{r}_0+k\cdot \boldsymbol{v}_1|-1\leq k\leq 1\}$. Note that here we can directly perform DFT on the corresponding line to obtain the precise spectrum.

Figure~\ref{fig:OnOffManifold} presents the on-/off- manifold spectra and the generalization accuracy. As shown in Figure~\ref{fig:OnManifold}, the on-manifold spectrum keeps introducing high-frequency components to memorize the perturbed point, which is not the case for the off-manifold spectrum. Figure~\ref{fig:OffManifold} shows that after the perturbed point is memorized, the off-manifold spectrum is more biased towards the low-frequency components, resulting in a much flatter off-manifold prediction surface. The regularized prediction surface has low complexity and potentially improves the model's generalization performance on some test points which are not covered by the training data manifold, leading to the second descent of the test error. As illustrated in Figure~\ref{fig:OnOffACC}, the on-manifold accuracy slightly decreases after memorizing the perturbed point, whereas the off-manifold accuracy keeps increasing in the same period.

\section{Monitor the Test Curve without Test Set}

In Figure~\ref{fig:FreqAndError}, we have seen that the peak of $R_k$ seems to synchronize with the start of the second descent of the test error. Based on this observation, we show that it may be possible to monitor the second descent of test error without any test set.

\subsection{Consistency of $R_k$ on Training and Test Sets}

One obstacle of using training errors to monitor test ones is that they do not vary consistently during training, especially when the number of epochs is large. This inconsistency is a result of the direct involvement of training errors in the optimization function, whereas test errors are excluded. Therefore, if we want to use a metric calculated on the training set to monitor the behaviors of the test curve, the most essential step is to cut off its connection to the optimization function.

Apparently, $R_k$, which measures the local variation of DNNs, satisfies this requirement. More importantly, $R_k$ is calculated without any label information, which further weakens its links to the optimization function. To verify its consistency on the training and test sets, we calculated $R_k$ on the test set, which is presented in Appendix~B and shows no significant difference with $R_k$ calculated on the training set (see Figure~\ref{fig:FreqAndError}). To compare the consistency of training and testing curves for errors and $R_k$, we calculated their short-time Pearson correlation coefficients (PCCs), whose value at training epoch $t$ is the PCC in a sliding window of length $l$ (set to 100 in our experiments) at $t$:
\begin{align}
p_l(t;\boldsymbol{x},\boldsymbol{y}) = \frac{\sum_{i=t}^{t+l-1}(\boldsymbol{x}_i-\bar{\boldsymbol{x}}_t)(\boldsymbol{y}_i-\bar{\boldsymbol{y}}_t)}{\sqrt{\sum_{i=t}^{t+l-1}(\boldsymbol{x}_i-\bar{\boldsymbol{x}}_t)^2\sum_{i=t}^{t+l-1}(\boldsymbol{y}_i-\bar{\boldsymbol{y}}_t)^2}},
\end{align}
where $\bar{\boldsymbol{x}}_t=\frac{1}{l}\sum_{j=t}^{t+l-1}\boldsymbol{x}_j$ and $\bar{\boldsymbol{y}}_t=\frac{1}{l}\sum_{j=t}^{t+l-1}\boldsymbol{y}_j$.

\begin{figure}[ht]\centering
\includegraphics[width=.85\columnwidth,clip]{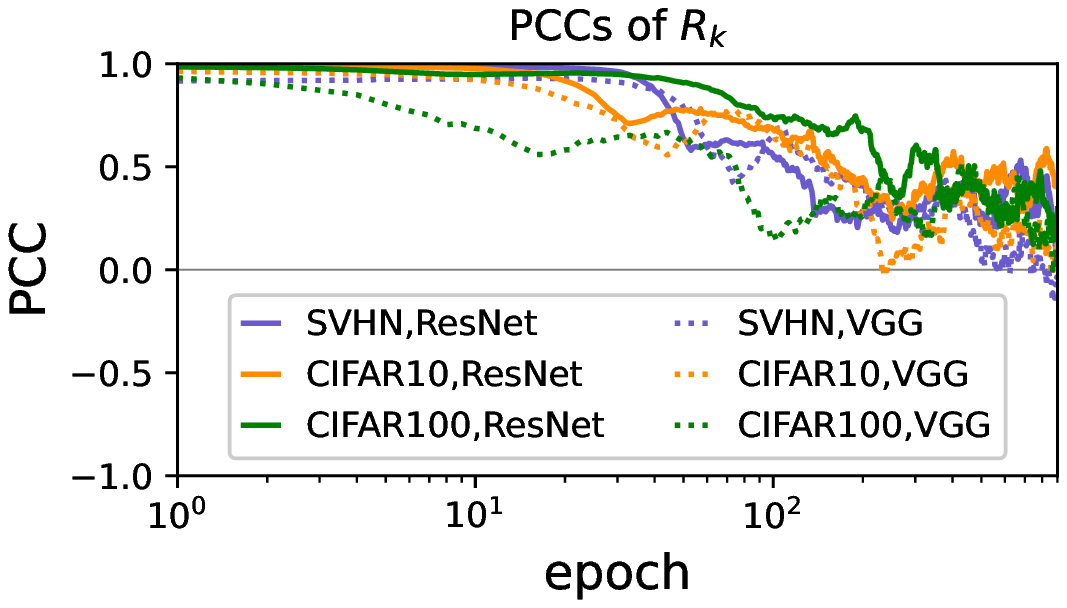}
\includegraphics[width=.85\columnwidth,clip]{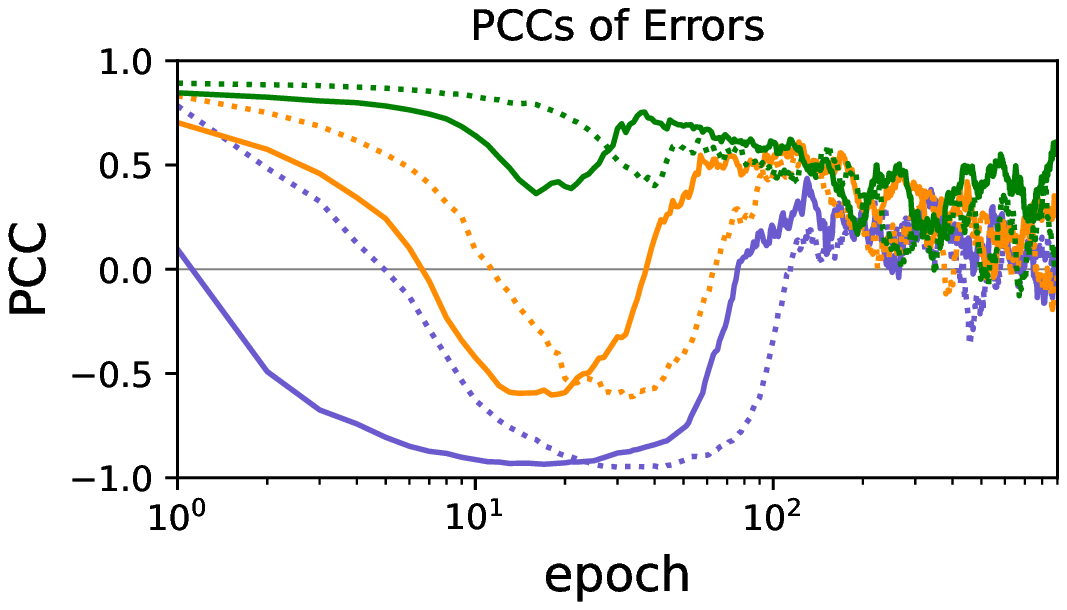}
\caption{Short-time PCCs of $R_k$ and errors. The colors represent different datasets, whereas the styles of lines indicate different categories of architectures. The short-time PCC of $R_k$ is averaged over $k=1,2...,64$. The horizontal axis is shown in logarithmic scale.} \label{fig:PCCt}
\end{figure}

Figure~\ref{fig:PCCt} shows short-time PCCs of the training and test errors, and $R_k$ on the training and test sets as well. Compared with the error, $R_k$ shows better consistency on the training and testing sets. Observe that the short-time PCC decreases when the number of training epochs is very large, because in the late stage, the variation tendency of the errors or $R_k$ is so slow that the noise of variation dominates the short-time PCCs instead of the overall variation tendency.

From the above observation and the discussions in the last section, we can conclude that the local variation of DNNs seems to have a consistent pattern in the input space during training, which makes $R_k$ an excellent metric to indicate the training progress.

\subsection{Use $R_k$ to Monitor the Second Descent of the Test Error}

We have verified the consistency of $R_k$ on training and test sets. To monitor the test curves, the metric should also have nontrivial meaningful connections to the test behaviors of DNNs. In Figure~\ref{fig:FreqAndError}, we can clearly observe the synchronization of the peak of $R_k$, which is calculated on the training set, with the start of the second descent of the test error. This subsection further explores this connection.

Let $R_{k,t}$ denote $R_k$ at the $t$-th epoch ($R_{k, t}$ was smoothed by mean value filtering with a window length of 10 epochs), and $E_{t}$ the corresponding test error. Similar to early stopping, we searched the peak of $E_t$ and $R_t=\sum_k \alpha_k R_{k,t}$ with the patience of 30 epochs, where $\alpha_k$ is the weight of $R_{k,t}$ and all set to $1$ in our experiments. More specifically, we performed early stopping twice: first, we tried to find the number of epochs for minimal $R_t$ and $E_t$, denoted by $T_{R,\min}$ and $T_{E,\min}$, respectively; then, we used $T_{R,\min}$ and $T_{E,\min}$ as start points to search for the epochs of maximal $R_t$ and $E_t$, denoted by $T_{R, \text{peak}}$ and $T_{E, \text{peak}}$, respectively.

\begin{figure*}[ht]\centering
\subfigure[]{\label{fig:SecondDescent}\includegraphics[width=.3\textwidth,clip]{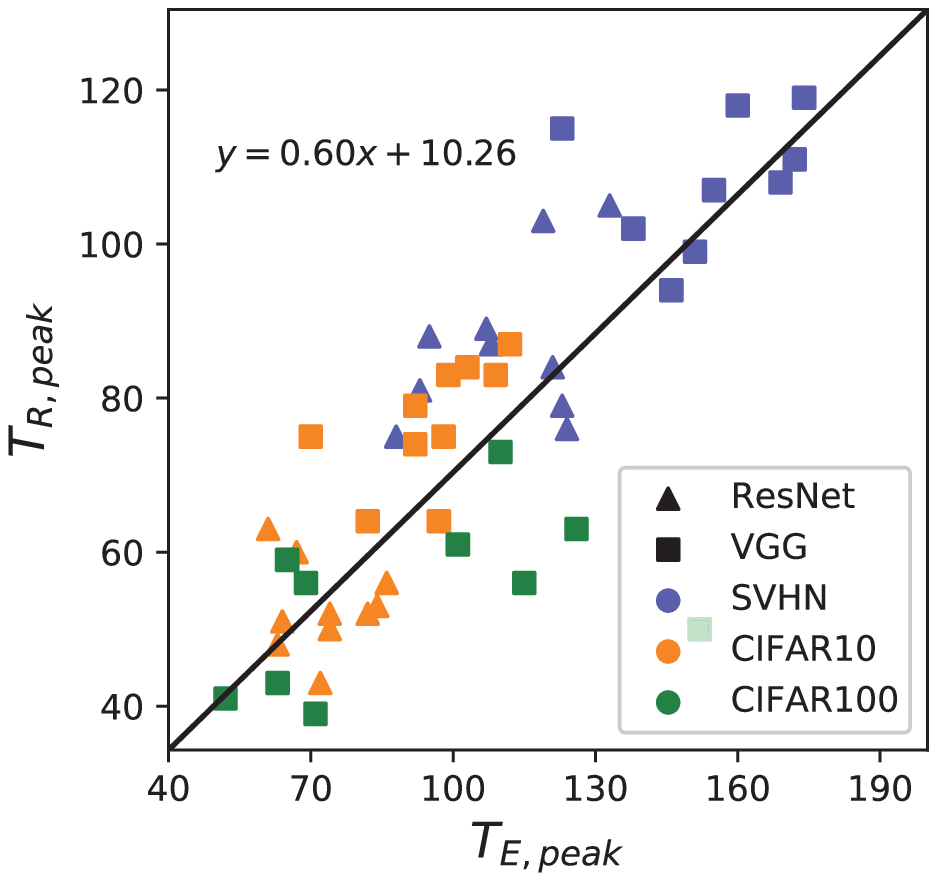}}
\subfigure[]{\label{fig:MaxRate}\includegraphics[width=.3\textwidth,clip]{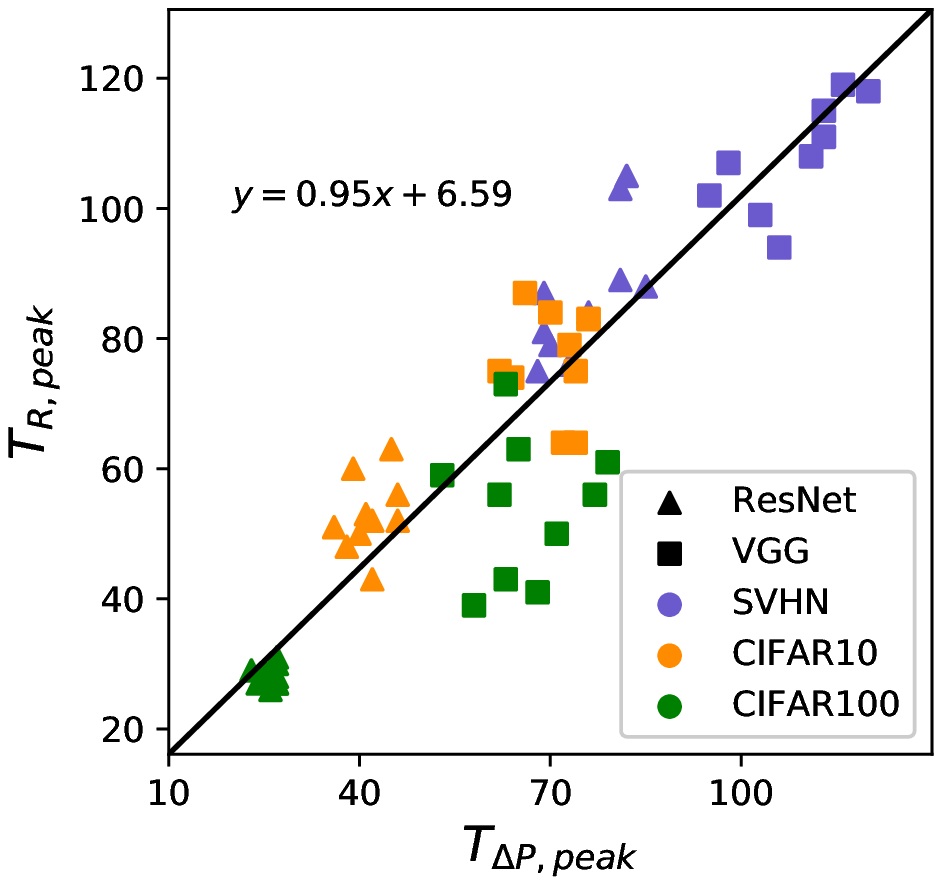}}
\subfigure[]{\label{fig:EarlyStop}\includegraphics[width=.3\textwidth,clip]{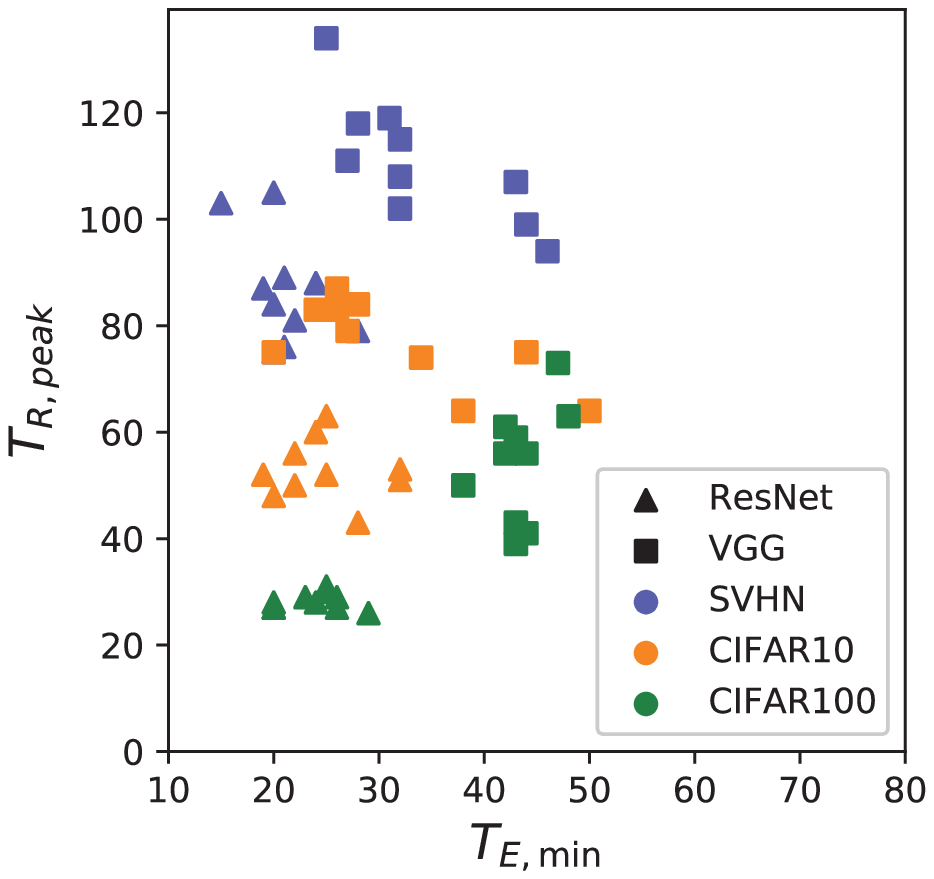}}
\caption{(a) $T_{R,\text{peak}}$ w.r.t. $T_{E,\text{peak}}$; (b) $T_{R,\text{peak}}$ w.r.t. $T_{\Delta P,\text{peak}}$; (c) $T_{R,\text{peak}}$ w.r.t. $T_{E,\min}$. The colors represent different datasets, whereas the shapes indicate different categories of DNNs, e.g., green squares represent VGG on CIFAR100. The black line is fitted by linear regression. Each experiment was repeated five times. The Pearson correlation coefficients for (a) and (b) are 0.88 and 0.92, respectively.}
\end{figure*}

We trained different models on several datasets (SVHN: ResNet18 and VGG11; CIFAR10: ResNet18 and VGG13; CIFAR100: ResNet34 and VGG16) with different levels of label noise ($10\%$ and $20\%$) for five runs, and explored the relationship between the spectra and the test behaviors. As shown in Figure~\ref{fig:SecondDescent}, despite of different models and datasets, we can clearly observe a linear positive correlation between $T_{R, \text{peak}}$ and $T_{E, \text{peak}}$, suggesting that it is possible to predict the second descent of the test error with only the training set. Moreover, the spectrum can also indicate how the peak of the double descent moves when the model width varies, which further verifies the connection between the spectrum and the epoch-wise double descent (see Appendix~C for more details).

$T_{R, \text{peak}}$ is also related to the decreasing rate of errors on the perturbed set. Let $P_{t}$ denote the perturbed error at the $t$-th epoch, and $\Delta P_{t}=-(P_{t}-P_{t-1})$ the decreasing rate. We searched the peak of $\overline{\Delta P}_t=\frac{1}{2\Delta T+1}\sum_{i=-\Delta T}^{\Delta T}\Delta P_{t+i}$ ($\Delta T=5$ in our experiments) due to the large variance of $\Delta P_{t}$, and the corresponding epoch was denoted by $T_{\Delta P,\text{peak}}$. Figure~\ref{fig:MaxRate} shows that when high-frequency components reach their largest ratio, the perturbed error decreases the fastest, suggesting that the spectrum can also be applied to studying some more subtle behaviors.

\subsection{Discussion}

Overfitting is another important test behavior. A validation set is usually reserved to indicate when overfitting happens. However, this may reduce the training performance because of the reduced training set size. A more popular approach is that after obtaining a suitable number of training epochs from the validation set, we combine the training set and the validation set to train the model for the same number of epochs. However, this approach also has two shortcomings: 1) there is no guarantee that the ``sweet spot'' of the training epoch does not change when the training and validation sets are combined; and, 2) it is time-consuming to train the model several times. If we can discover a novel metric which can be calculated on the training set but effective enough to predict the epoch of overfitting (just like the spectrum to the second descent of the test error), these problems can be easily solved. However, we are not able to find a direct connection between $T_{R,\text{peak}}$ and $T_{E,\min}$ [see Figure~\ref{fig:EarlyStop}]. It is also one of our future research directions to find a metric to indicate $T_{E,\min}$.

\section{Related Work} \label{sect:related}

There are many studies related to the themes of this paper:
\paragraph{Generalization and Memorization.}
Over-parameterized DNNs are believed to have large expressivity, usually measured by the number of linear regions in the input space~\cite{Pascanu2014,Montufar2014,Poole2016,Arora2018,Zhang2020}. However, it cannot explain why a DNN, whose capacity is large enough to fit random noise, still has low variance on normal datasets~\cite{zhang2017}. Arpit \emph{et al.}~\shortcite{Arpit2017} examined the role of memorization in deep learning and showed its connections to the model capacity and generalization. Zhang \emph{et al.}~\shortcite{Zhang2020a} studied the interplay between memorization and generalization and empirically showed that different architectures exhibit different inductive biases. Despite of these studies, memorization and generalization of DNNs is still an open problem requiring more exploration.

\paragraph{Learning Bias.} Learning bias suggests that SGD has an implicit inductive bias when searching for the solutions, e.g., from learning patterns to memorizing noise~\cite{Arpit2017}, from simple to complex~\cite{Kalimeris2019}, or from low frequency to high frequency~\cite{Rahaman2019,Xu2019,Xu2019a}. For spectral bias, some studies investigated the convergence rate of different frequency components, but they were very specific, e.g., using DNNs with infinity width or synthetic datasets~\cite{Ronen2019,Cao2019}. Moreover, they all suggested that learning bias is monotonic, which is different from our findings.

\paragraph{Double Descent.}
There is epoch-wise double descent and model-wise double descent~\cite{Nakkiran2020}. The former is a general phenomenon observed by many studies~\cite{Belkin2018,Geiger2020,Yang2020}, whereas the latter was proposed recently, inspired by a unified notion of ``Effective Model Complexity''~\cite{Nakkiran2020}. Our work provides a novel perspective to analyze the epoch-wise double descent.

\section{Conclusions} \label{sect:conclusion}

Our research suggests that we need to rethink the connections among generalization, memorization and the spectral bias of DNNs. We studied the frequency components of DNNs in the data point neighbors via Fourier analysis. We showed that the monotonicity of the spectral bias does not always hold, because the off-manifold prediction surface may reduce its high-frequency components in the late training stage. Though perturbed points on the training data manifold remain memorized by the on-manifold prediction surface, this implicit regularization on the off-manifold prediction surface can still help improve the generalization performance. We further illustrated that unlike errors, the spectrum shows remarkable consistency on the training sets and test sets. Based on these observation, we found the potential correlation of the spectrum, calculated on the training set, to the second descent of the test error, suggesting that it may be possible to monitor the test behavior using the training set only.

Our future research will: a) analyze the spectrum of DNNs in other learning tasks, e.g., natural language processing, speech recognition and so on; b) find a new metric which can be easily calculated on the training set, but effective enough to indicate the start of overfitting; c) explore the role that SGD and skip connection play in the spectra of DNNs.

\section*{Acknowledgements}

This work was supported by the CCF-BAIDU Open Fund under Grant OF2020006, the Technology Innovation Project of Hubei Province of China under Grant 2019AEA171, the National Natural Science Foundation of China under Grants 61873321 and U1913207, and the International Science and Technology Cooperation Program of China under Grant 2017YFE0128300.

\bibliographystyle{named}
\bibliography{xiaozhangbib}

\clearpage
\onecolumn
\appendix
\section{$R_k$ with Different Levels of Label Noise}
\begin{figure*}[!ht]\centering
\subfigure[SVHN]{
    \begin{minipage}[b]{0.31\textwidth}
    \includegraphics[width=\textwidth]{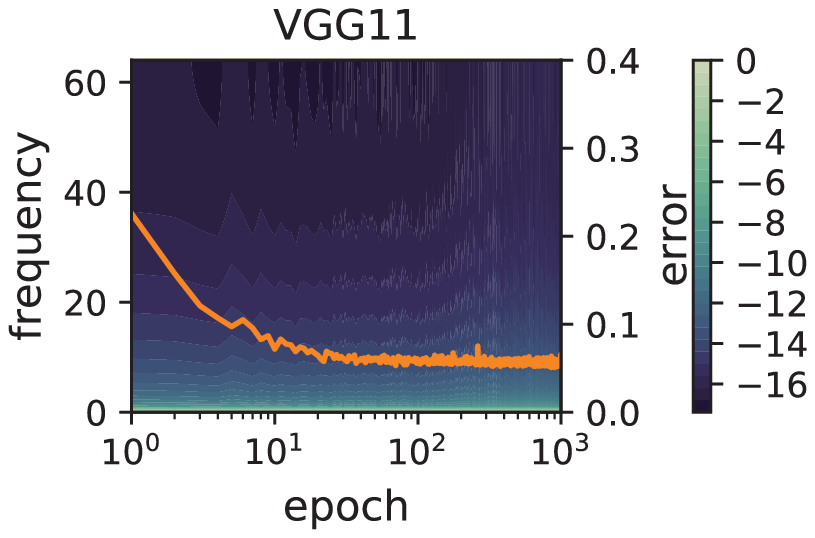} \\
    \includegraphics[width=\textwidth]{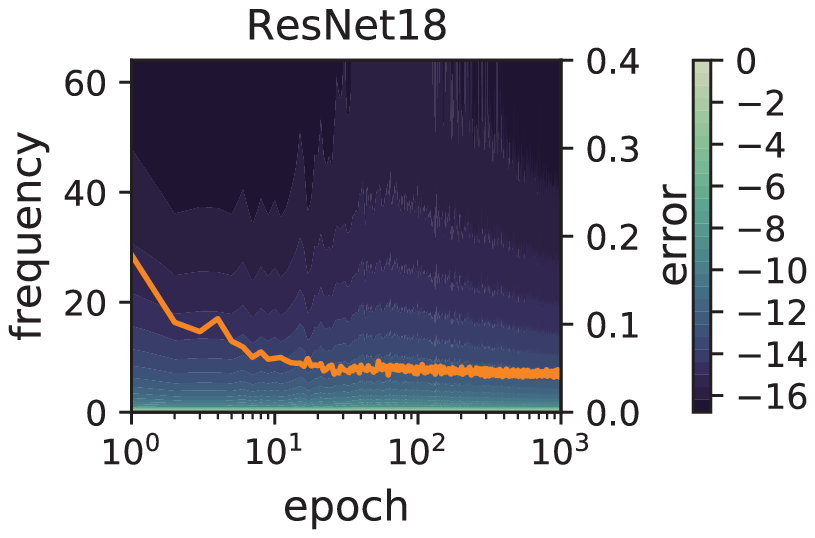}
    \end{minipage}
}
\subfigure[CIFAR10]{
    \begin{minipage}[b]{0.31\textwidth}
    \includegraphics[width=\textwidth]{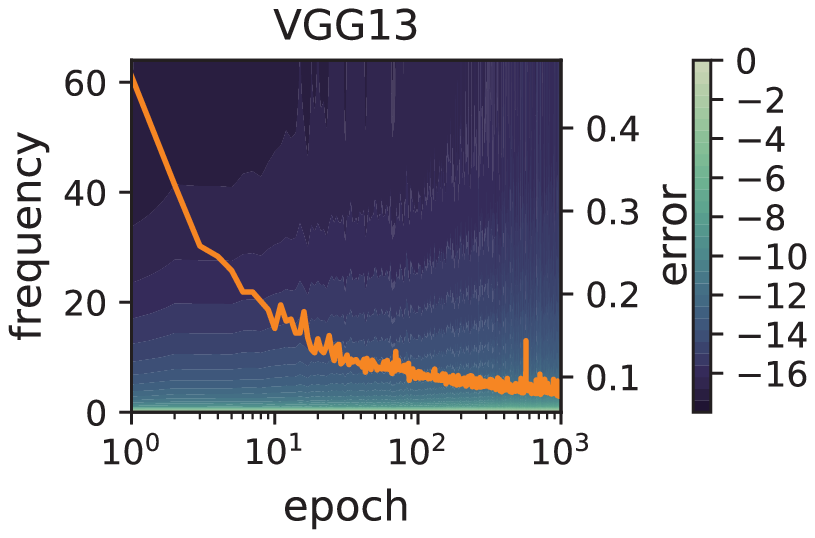} \\
    \includegraphics[width=\textwidth]{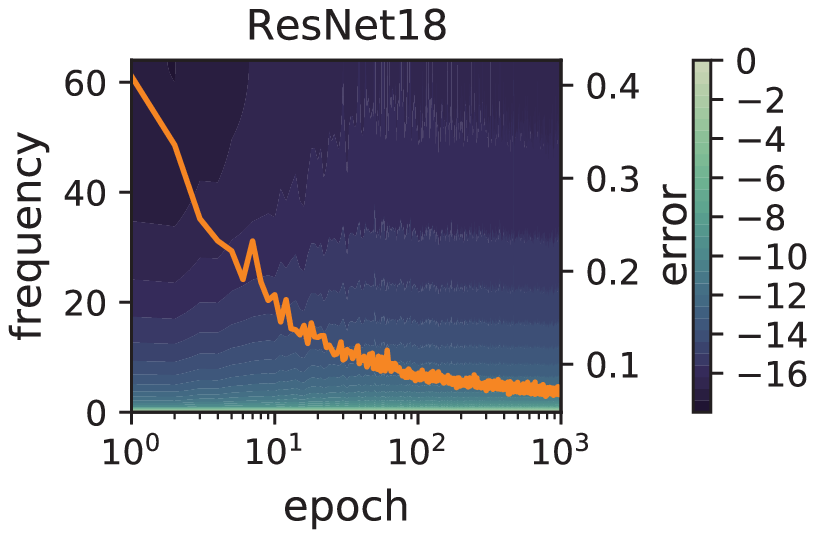}
    \end{minipage}
}
\subfigure[CIFAR100]{
    \begin{minipage}[b]{0.31\textwidth}
    \includegraphics[width=\textwidth]{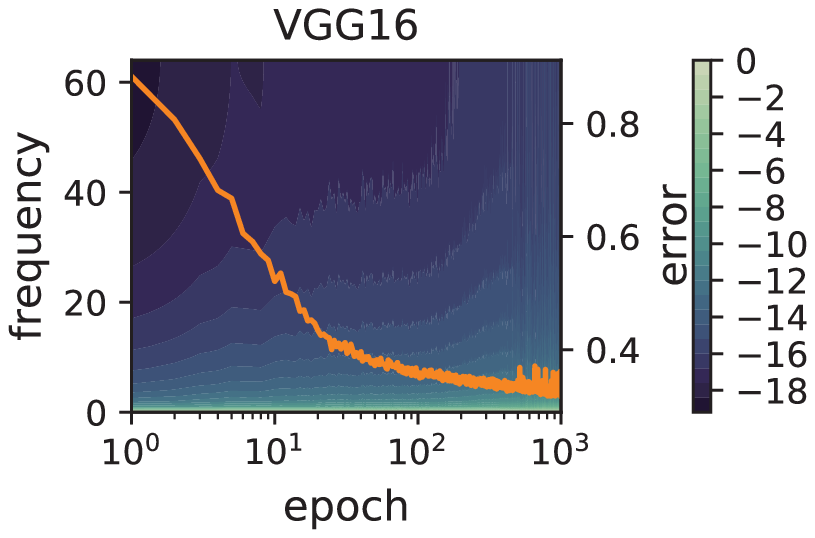} \\
    \includegraphics[width=\textwidth]{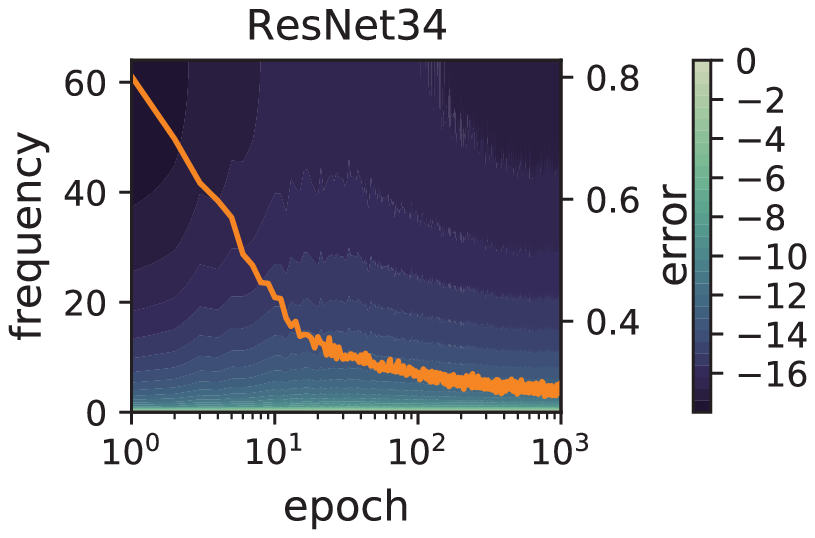}
    \end{minipage}
}
\caption{Models trained on SVHN, CIFAR10 and CIFAR100 with \textbf{0\%} label noise. The heat map with contour lines depicts $R_k$ which is calculated on the training set, where $k$ denotes the ``frequency''.}
\end{figure*}
\begin{figure*}[!ht]\centering
\subfigure[SVHN]{
    \begin{minipage}[b]{0.31\textwidth}
    \includegraphics[width=\textwidth]{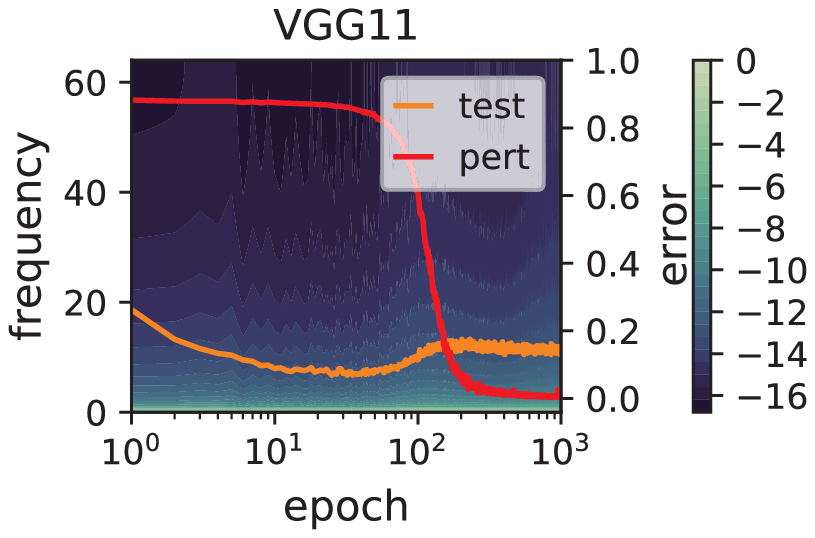} \\
    \includegraphics[width=\textwidth]{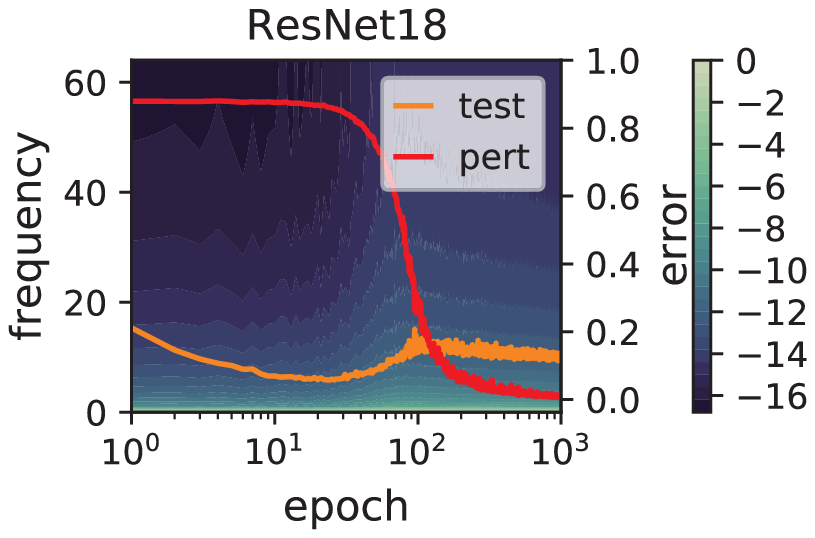}
    \end{minipage}
}
\subfigure[CIFAR10]{
    \begin{minipage}[b]{0.31\textwidth}
    \includegraphics[width=\textwidth]{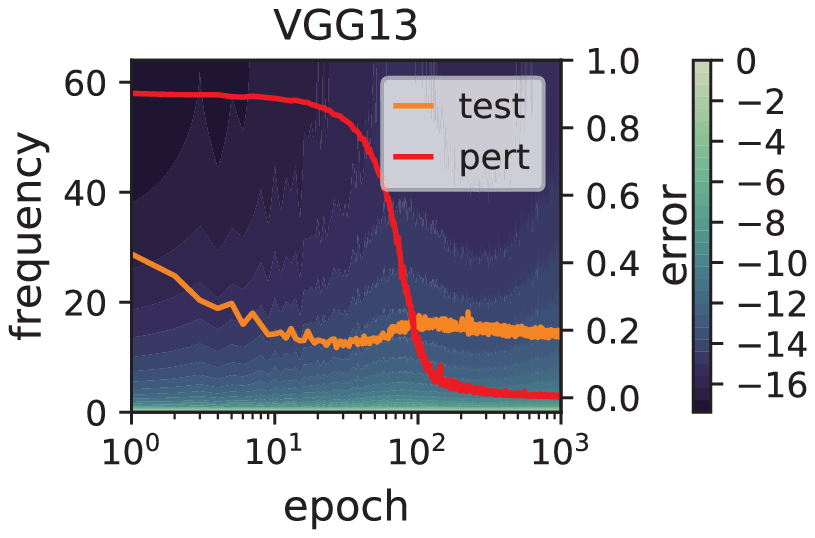} \\
    \includegraphics[width=\textwidth]{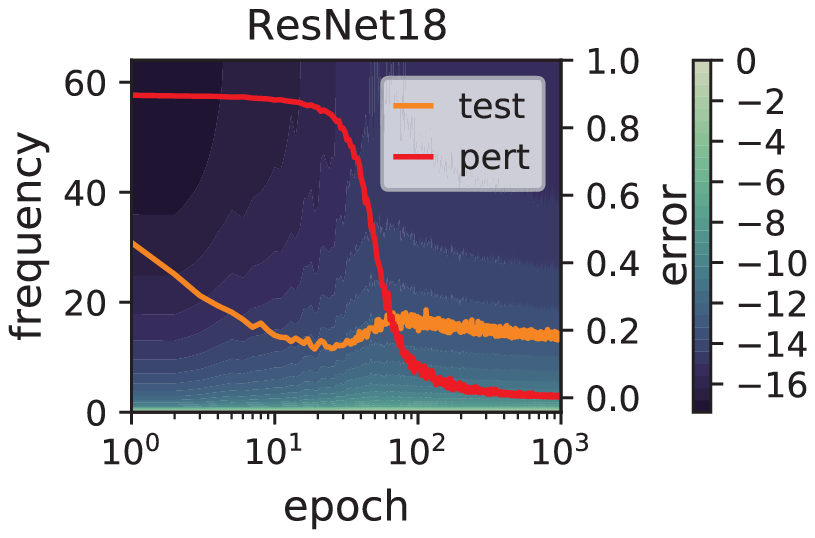}
    \end{minipage}
}
\subfigure[CIFAR100]{
    \begin{minipage}[b]{0.31\textwidth}
    \includegraphics[width=\textwidth]{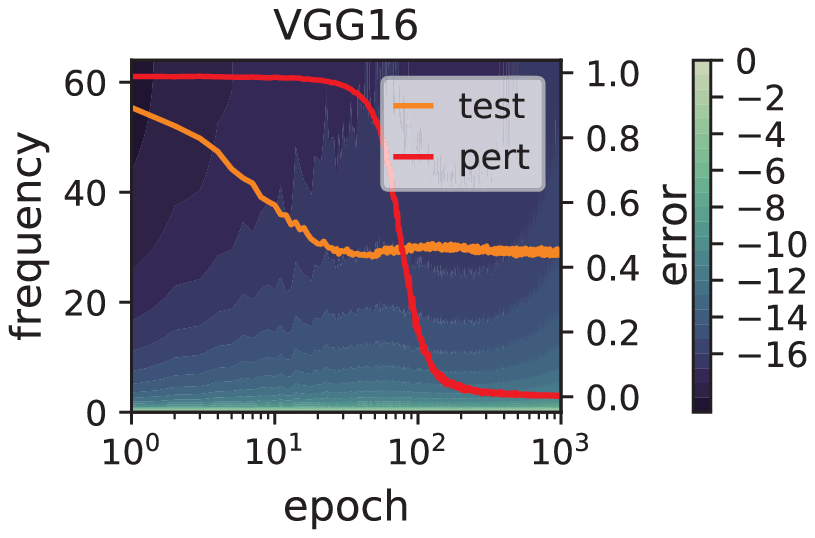} \\
    \includegraphics[width=\textwidth]{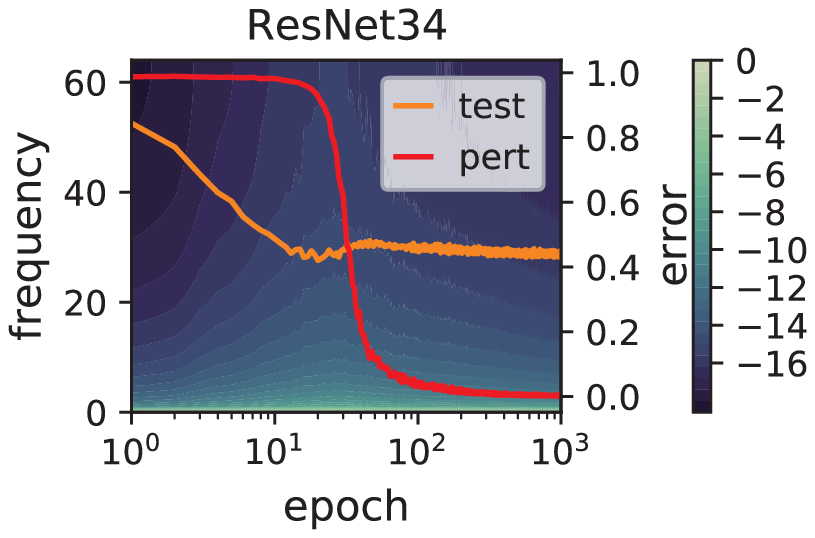}
    \end{minipage}
}
\caption{Models trained on SVHN, CIFAR10 and CIFAR100 with \textbf{20\%} label noise. The heat map with contour lines depicts $R_k$ which is calculated on the training set, where $k$ denotes the ``frequency''.}
\end{figure*}

\clearpage
\section{$R_k$ on Test Sets}

\begin{figure*}[ht]\centering
\subfigure[SVHN]{
    \begin{minipage}[b]{0.31\textwidth}
    \includegraphics[width=\textwidth]{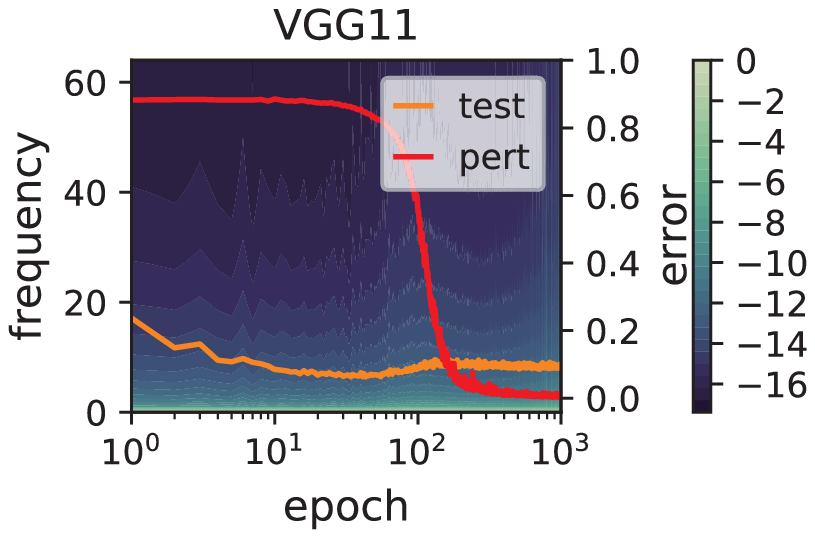} \\
    \includegraphics[width=\textwidth]{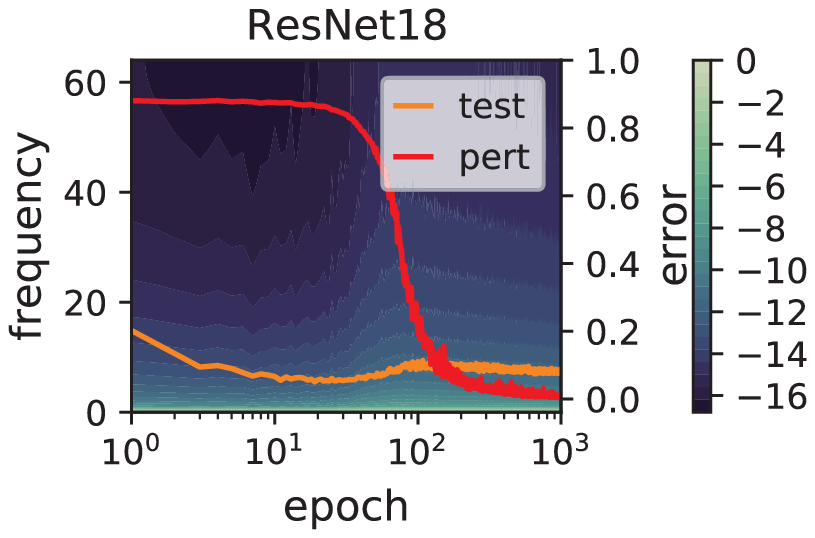}
    \end{minipage}
}
\subfigure[CIFAR10]{
    \begin{minipage}[b]{0.31\textwidth}
    \includegraphics[width=\textwidth]{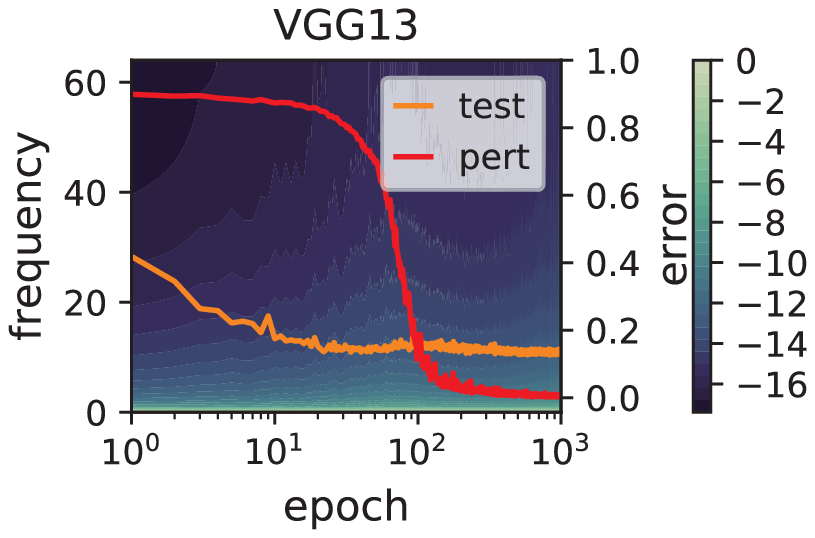} \\
    \includegraphics[width=\textwidth]{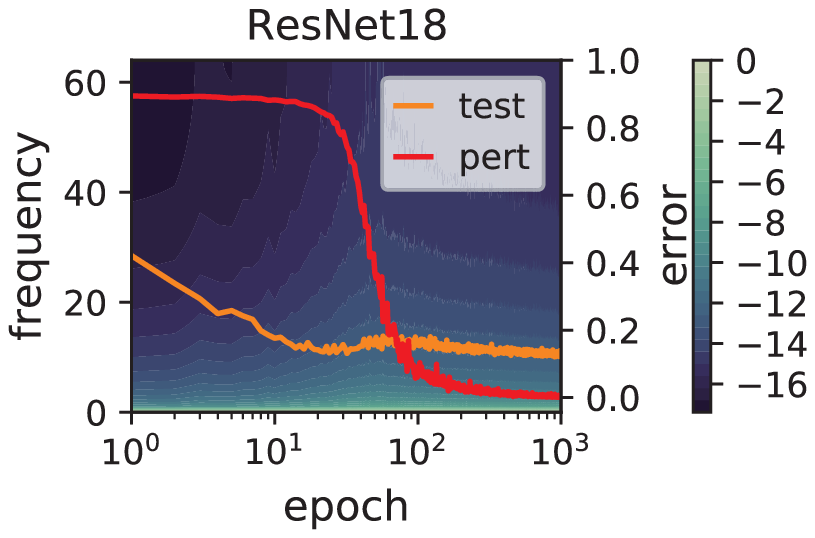}
    \end{minipage}
}
\subfigure[CIFAR100]{
    \begin{minipage}[b]{0.31\textwidth}
    \includegraphics[width=\textwidth]{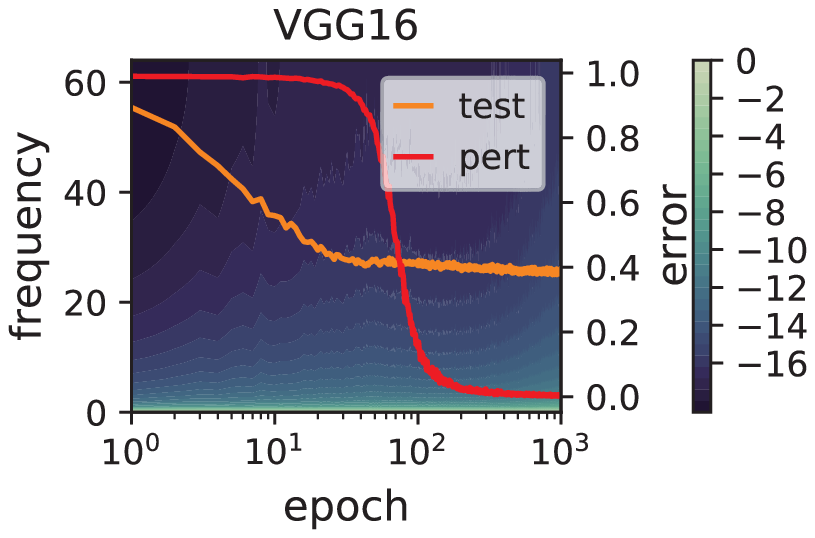} \\
    \includegraphics[width=\textwidth]{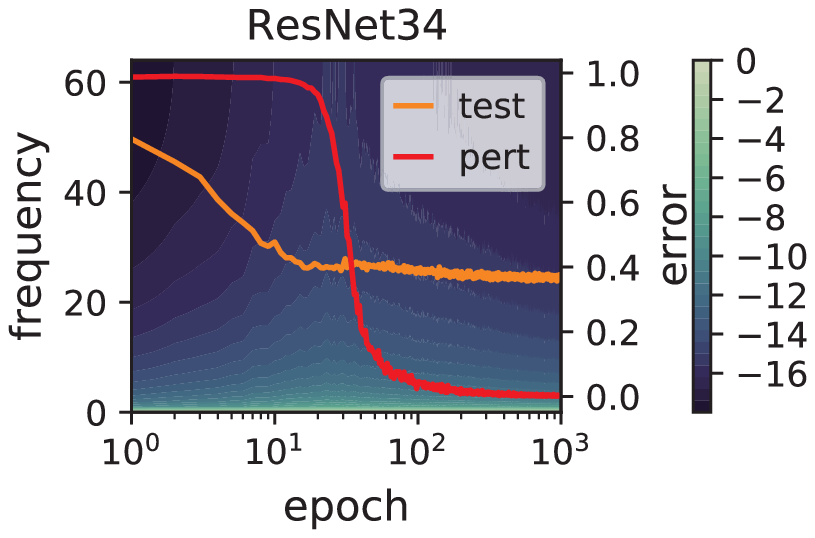}
    \end{minipage}
}
\caption{Models trained on SVHN, CIFAR10 and CIFAR100 with 10\% label noise. The heat map with contour lines depicts $R_k$, which is calculated on the \textbf{test set}, where $k$ denotes the ``frequency''.}
\end{figure*}

\section{$R_k$ v.s. Model Width}

We show that the peak of the spectrum moves along with the peak of the test error when the width of DNN varies. We trained ResNet18 with different width on CIFAR10 for 200 epochs, using Adam optimizer with learning rate 0.0001. For each convolutional layer, we set the number of filters $q\in[0.5, 2]$ times the number of filters in the original model. We then examined the synchronization between the peaks of the spectrum and the test error.

\begin{figure*}[ht]\centering
\subfigure[Test errors]{
    \includegraphics[width=.31\textwidth]{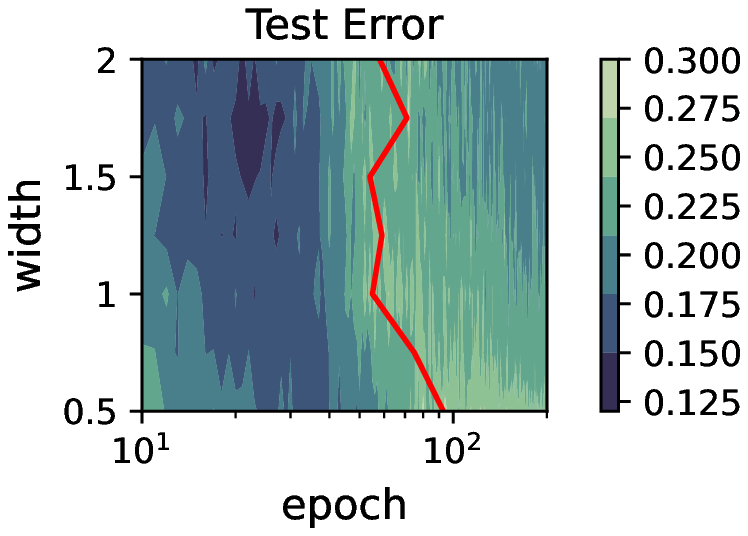}
}
\subfigure[$R_{32}$]{
    \includegraphics[width=.31\textwidth]{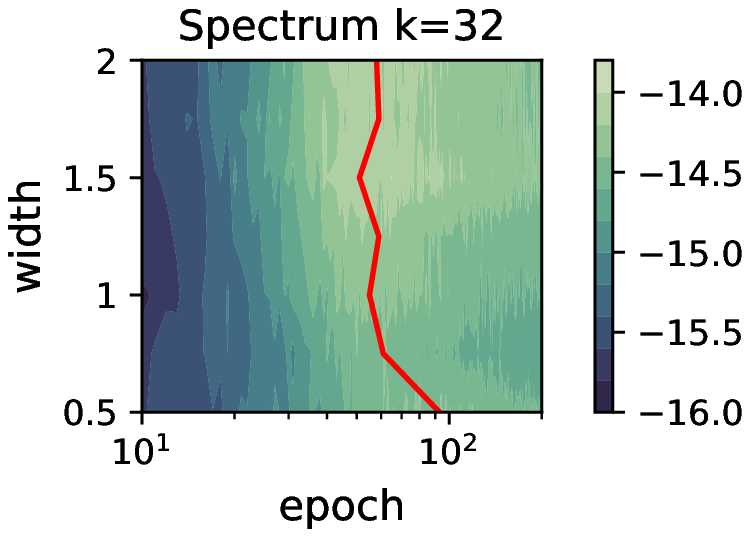}
}
\subfigure[$R_{63}$]{
    \includegraphics[width=.31\textwidth]{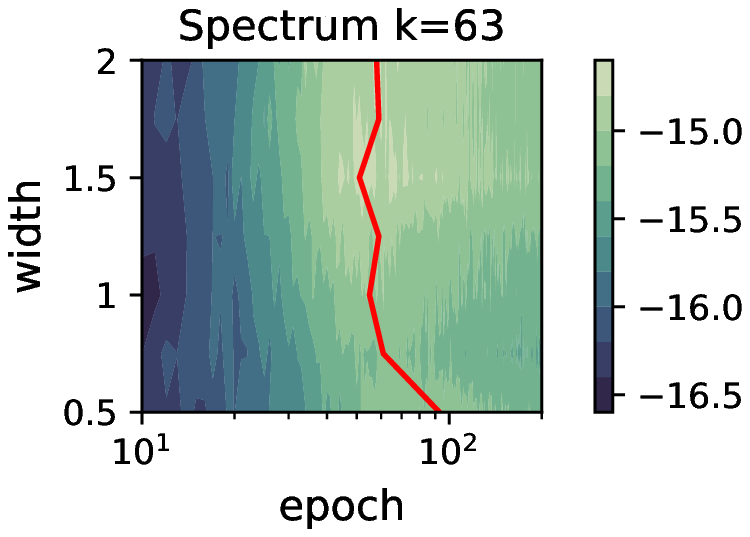}
}
\caption{Test error and $R_k$ w.r.t. model width. The original model is ResNet18 trained on CIFAR10 with 20\% label noise. The red lines depict the exact positions of peaks of test errors and $R_k$.} \label{fig:width}
\end{figure*}

The results are shown in Figure~\ref{fig:width}. First, we can observe that when the model becomes wider, the prediction surface is allowed to introduce more high-frequency components because of the larger model complexity; Second, as illustrated by the red lines in Figure~\ref{fig:width}, the peak of the spectrum moves along with the peak of the test error, which further verifies the connection between the spectrum and the generalization behavior of DNNs.

\end{document}